\documentclass[runningheads]{llncs}
\usepackage{graphicx}
\usepackage{cite}
\usepackage{todonotes}

\usepackage{booktabs}
\usepackage{multirow}

\newcommand{\repeatthanks}{\textsuperscript{\thefootnote}}
\begin{document}
\title{Knowledge Graph Induction enabling Recommending and Trend Analysis:\\A Corporate Research Community Use Case}
\titlerunning{Exploiting Induced Knowledge Graphs for Recommending and Analytics}
%
\author{Nandana Mihindukulasooriya\thanks{Equal contributions} \and
Mike Sava\repeatthanks \and
Gaetano Rossiello\repeatthanks \and 
Md Faisal Mahbub Chowdhury\repeatthanks \and 
Irene Yachbes \and
Aditya Gidh \and
Jillian Duckwitz \and 
Kovit Nisar \and 
Michael Santos \and
Alfio Gliozzo
}

\authorrunning{Mihindukulasooriya et al.}
%
\institute{IBM Research AI, Yorktown Heights, NY, USA}
%
\maketitle              
\begin{abstract}
A research division plays an important role of driving innovation in an organization. Drawing insights, following trends, keeping abreast of new research, and formulating strategies are increasingly becoming more challenging for both researchers and executives as the amount of information grows in both velocity and volume.  In this paper we present a use case of how a corporate research community, IBM Research, utilizes Semantic Web technologies to induce a unified Knowledge Graph from both structured and textual data obtained by integrating various applications used by the community related to research projects, academic papers, datasets, achievements and recognition. In order to make the Knowledge Graph more accessible to application developers, we identified a set of common patterns for exploiting the induced knowledge and exposed them as APIs. Those patterns were born out of user research which identified the most valuable use cases or user pain points to be alleviated. We outline two distinct scenarios: recommendation and analytics for business use.  We will discuss these scenarios in detail and provide an empirical evaluation on entity recommendation specifically. The methodology used and the lessons learned from this work can be applied to other organizations facing similar challenges. 

\end{abstract}

\section{Introduction}

Research and innovation is the heart of any organization that is focused on advancing technologies to meet the challenges of solving real world problems by bridging the business needs with scientific discoveries. In fast moving research areas such as artificial intelligence or quantum computing, there is a tremendous growth of research activities in both velocity and volume happening within and outside the organization~\cite{savage2020race, bornmann2015growth}. It is challenging to understand the trends and draw insights, and doing so manually is becoming unfeasible. Nevertheless, such insights are of utmost important for the executives who make strategy decisions on the impact of current investments and decide on future directions~\cite{salatino2021detection} and for the researchers who are looking for effective collaborations to optimize the reuse of research assets. In addition, in large organizations involving thousands of people and various scientific disciplines, it is difficult to keep abreast of individual projects. Weekly updates are often overwhelming but essential to make sure that people are informed of progress, to prevent redundant work, enhance re-usability, and cross fertilize ideas and assets. However, those has to be personalized to each person's user's interests to keep the information overload minimal. 

One major challenge in generating insights is that generally data is scattered across different applications in their own siloed spaces. If integrated manually, this requires a lot of effort and hinders their full potential use for downstream applications. Thus, it is useful for an organization to have a unified integrated view of the data. Furthermore, these applications capture both structured metadata and also a lot of unstructured textual data. It's challenging to analyze the useful insights hidden in large volumes of text and uncover the insights. 

For example, in the IBM research community, there are different applications for  managing research projects, academic papers, datasets, internal achievements and external recognition. Researchers are both the content providers who contribute to these applications as well as end users that gets the recomendations and insights. From the adoption point of view, it is important that they have to spend only a minimum amount of valuable time without duplication of effort in multiple apps for the same information and get high value and useful insights in order to increase the engagement.

Before jumping to the solutions, we have first conducted a user study to understand the most valuable user pain points to be alleviated. Through a set of in-depth interviews from a set of selected users in different stages of their career, recommendations and trend analytics were identified as two main use cases that most requested by the community, as discussed in Section \ref{sec:use_case}. 

The aforementioned scenario provided us an excellent use case to test the boundaries of Knowledge Graph Induction (KGI) framework which is presented in this paper. Specifically, we apply our technology to mitigate some of the challenges in a corporate research community: IBM Research. While we restrict our focus to a research community in this paper, KGI framework can be applied to any organization that has a large volume of structured and unstructured data to be integrated and analyzed.



We will discuss how we address the common challenges of extraction of knowledge from both structured and unstructured data, how to enrich the KG from information available in the vast amounts of unstructured text 
and how to use the enriched KG to power Knowledge Exploitation Patterns (KEP) for entity recommendation and trend analytics. We will also discuss how the external encyclopedic knowledge such as Wikidata~\cite{vrandevcic2014wikidata} can be seamlessly integrated to internal knowledge enabling traversal following the Linked Data principles to get more context or provide more structure to the data using the taxonomic knowledge.


The main contributions of this paper are as follows:

\begin{itemize}
  
  \item We introduce an end-to-end framework for Knowledge Graph Induction from both structured, semi-structured, and unstructured data. KGI is easily portable across domains and enables the reuse of high level abstractions, i.e. KEP, for recommending and trend analysis. 
  \item We introduce the KnowGL Parser, a Knowledge Generation and Linking approach based on transformer based generative models, which achieves the state of the art performances on information extraction benchmarks.
  \item We demonstrate the effectiveness of the KGI framework in two different scenarios: IBM research internal community and ISWC 2002-2021 proceedings.
  \item We discuss how a research organization can benefit from building a KG from both structured and unstructured data motivated by the pain points identified in a user study. 
\end{itemize}

The rest of this paper is structured as follows. Section \ref{sec:use_case} discusses the use cases identified after an extensive user study. Section \ref{KGI} introduces the KGI framework including knowledge integration, KnowGL Parser and evaluate the knowledge generation using an academic benchmark. Section \ref{sec:KEP} introduces KEP for Entity Recommendation, Trend Analysis, and Infobox Generation, providing and empirical evaluation of the recommending capabilities based on user evaluation. Section \ref{sec:relatedwork} presents a review of related work, while Section \ref{sec:conclusion} concludes the paper highlighting directions for future work. 
\section{Application Use Cases}
\label{sec:use_case}

The Apps@Research team, an application design and development team inside IBM Research, designs, develops, and supports a portfolio of cloud-based web applications providing rich, intuitive, integrated experiences that serve the unique needs of the IBM Research community. These include collaborative tools for:
\begin{itemize}
    \item proposing and reporting progress on research projects including tracking staff effort, milestones, and impact (Research Project Portal)
    \item tracking the status of papers submitted to conferences and journal throughout the cycle from submission to decision (Academic Paper Portal)
    \item cataloging datasets approved for use by the legal team and datasets published by our teams (Dataset catalog)
    \item nominating, reviewing and selecting projects to receive yearly internal accomplishment awards (Achievements Portal)
    \item tracking external recognition and awards won by IBM researchers (Recognition Portal)
\end{itemize}

The Apps@Research team engaged the IBM Research AI team to partner on ways to incorporate IBM Research's own artificial intelligence technologies to augment the user experience in these applications. The key motivations were to:

\begin{itemize}
    \item Unlock the content potential of the Apps@Research applications, which reflects the work and expertise across each division and teams. 
    \item Improve user experience by creating exceptional, well-curated, concise and personalized information.
    \item Leverage and offer a testbed for IBM Research’s own AI technology
\end{itemize}

In order to inform prioritization for the product roadmap for one of the most pervasive applications, we undertook a foundational user research study in 2020 to better understand user needs. The study included over 100 interviews and 220 survey responses from users of our applications. From this study, one key pain point was identified: because the content in our tools describe detailed research project proposals and plans of thousands of research projects, the content is too dense to be easily digestible. Users struggle in discovering relevant content and are under the perception that other users will find their content either. In turn, many users could become frustrated and stop using the tools for their key intended purposes - collaboration, innovation, and sharing updates. 

Our hypothesis was that if we were to find a way to help the content become more discoverable, personalized, and digestible, that users would be motivated to keep their content up-to-date and visit the tool more frequently to find synergies and sparking innovation across research projects. 

After doing some preliminary technical discovery and feasibility study with the AI Research team, we performed a more detailed user study. We recruited 12 participants from a representative sample of researcher and strategists at different stages in their career. They had varying experience with AI technology concepts. We conducted 60 minute structured interview sessions with users in which we asked open-ended questions and then engaged them in an interactive exercise in a mural application.





The purpose of the interactive exercise was to identify various possible use cases and to prioritize them. We gave the users a hypothetical ``\$100'' and asked them how they would ``spend'' the money, dividing among the use case ideas (Hundred dollar prioritization~\cite{khan2015comparison}). The purpose of the exercise was to understand the quantitative value that participants would ascribe to various use cases.

Upon completion of the interviews, we then performed a design thinking exercise called affinity mapping, to group ideas and identify common themes and patterns. We also analyzed the ``\$100  prioritization'' to help quantify the value of use cases to all the participants.

\begin{figure}[h!]
\includegraphics[width=\textwidth]{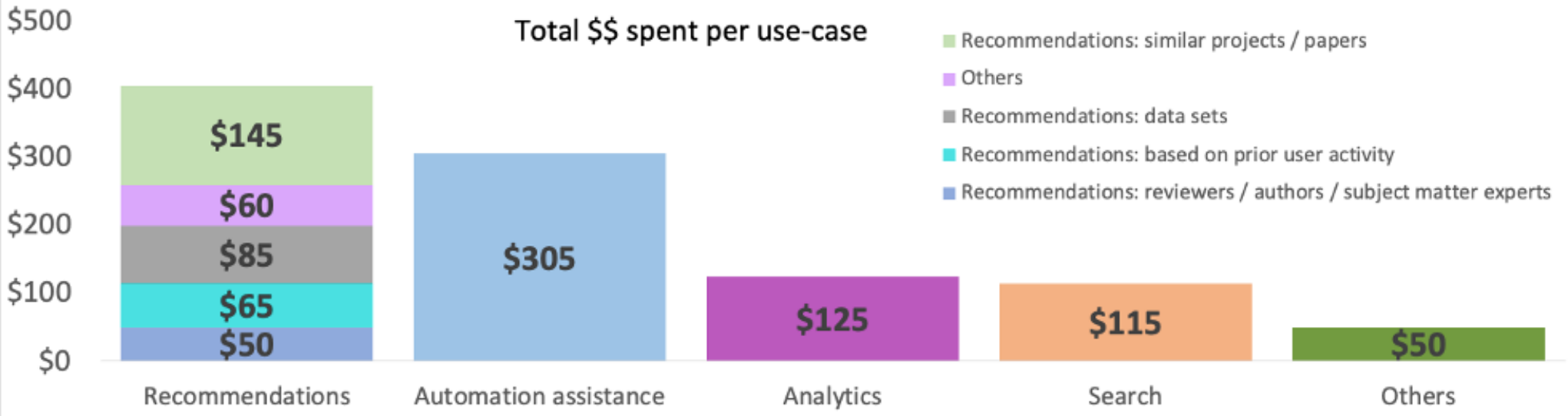}
\caption{User Interview \$100 ``spending'' results} \label{fig:user-spending}
\end{figure}


Figure~\ref{fig:user-spending} shows that most users identified ``recommendations'' as the most valuable use case. Recommendations would be automatically generated with information of which the user might otherwise not be aware. Recommendations would be personalized, based on users' previous activities such as papers/patents published, expertise, current projects, etc - all of which could be derived from data in our KG. Figure~\ref{fig:user-spending} also shows that users wanted several types of recommendations such as related research projects, relevant papers, collaborators, experts to review their papers, etc.



The second most valuable use case would be ``automation assistance''. This would include help to pre-fill forms in the various tools in a smart way, saving the user time and anticipating their needs. This was a technical requirements and having an integrated view in KG would allow us to pre-fill a lot of information in different applications based on the context. 

Next, users were interested in ``analytics'' - smart reports and dashboards that could be generated to provide business insights. The users have found that the data in our portals are dense and overwhelming and wanted to have high-level overview summaries so that can understand the common trends and dig more into the details.

Users were interested in improved Search and Filtering. Currently most of our applications' search is based on keywords and users were interested in more advanced semantic search capabilities. A KG would allow us to perform more complex structured searches.   




Knowing that recommendations ranked highest as the most important use case, we analyzed further which types of content would be of greatest interest, so that we could prioritize developing those features first. We found that users ascribed the most value to being recommended projects and papers.


The insights gained from the user study led us to focus on the following two use case scenarios:

\begin{itemize}
    \item \textbf{Recommending}: For researchers keeping abreast of colleagues' work (project status and publications) is very difficult in a large organization focusing on many technology areas.  This is a hindrance to effective collaborations and reuse of research assets. There is a need for technologies and tools to make this process more seamless. 
    \item \textbf{Trend Analysis}: For executives it is difficult to understand the breadth of the research portfolio, gain useful insights, and formulate a future strategy. There is a need to process large volumes of unstructured data and provide useful insights.
\end{itemize}

In the following sections, we will discuss our NLP and Semantic Web-driven approach for addressing these two main use case scenarios. 




\section{Knowledge Graph Induction}\label{KGI}

\begin{figure}[t]
\includegraphics[width=\textwidth]{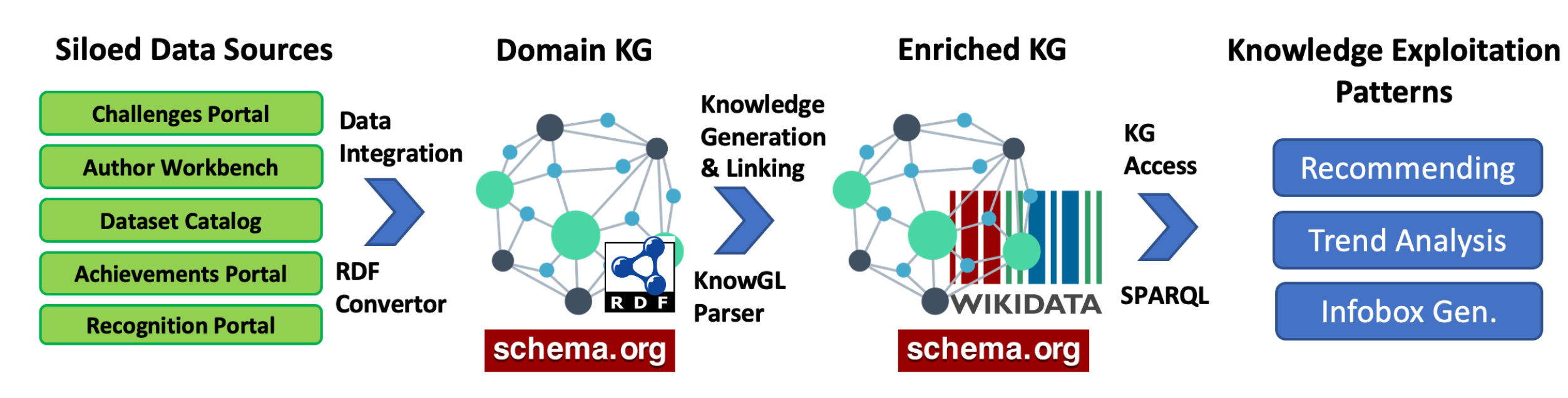}
\caption{The Knowledge Graph Induction framework} \label{fig:kg_structured}
\end{figure}

The overview of our KGI framework is illustrated in Figure \ref{fig:kg_structured}. It consists on three main conceptual blocks: data integration, whose main goal is to integrate heterogeneous semi-structured data from siloed applications using a domain ontology; knowledge extraction and linking, implemented by the \emph{KnowGL Parser}, a component capable of generating RDF compliant knowledge by processing all textual content attached to entities in the domain KG; and \emph{Knowledge Exploitation Patterns}, a set of abstractions over the induced KG that are domain-agnostic and generalized to use cases such as \emph{recommending} and \emph{analytics}. 

\subsection{Data Integration}

Data related to IBM Research is scattered across multiple siloed applications. We used a knowledge representation approach based on Semantic Web standards and unified them into a single KG with links to both internal entities as well as relevant information extracted from background knowledge sources such as Wikidata. 

Internal data pertains to items that are of particular interest to a research organization: research projects (science, strategy), people (eminence), academic publications and datasets (eminence), achievements and recognition (impact). 


Each of the applications provides an API to extract data, which is then processed through a RDF conversion pipeline following a process similar to RML-based tools~\cite{dimou2014rml}. For this purpose, a Research KG ontology was built by reusing and extending the Schema.org with classes and relations that were more specific to our use case. The data schema of each of the five applications were aligned to the ontology by a knowledge engineer and the mappings were created.

The Schema.org ontology was selected as the base because it covered most of the concepts in our applications and is used by some of our collaborators. In addition, entities and relations from  Wikidata are also reused. This enables us to easily integrate with third parties. The conversion process consists of (a) data extraction and (b) cleaning to normalize certain values, (c) mapping and RDF generation. Entity resolution is carried out to convert mentions to people, projects, and other entities to their canonical identifiers through a deterministic process. To this aim, we used unique identifiers such as emails and other internal conventions. 

\subsection{Knowledge Generation and Linking: KnowGL Parser}
\label{sec:wikidata_parser}

\begin{figure}[t]
\includegraphics[width=\textwidth]{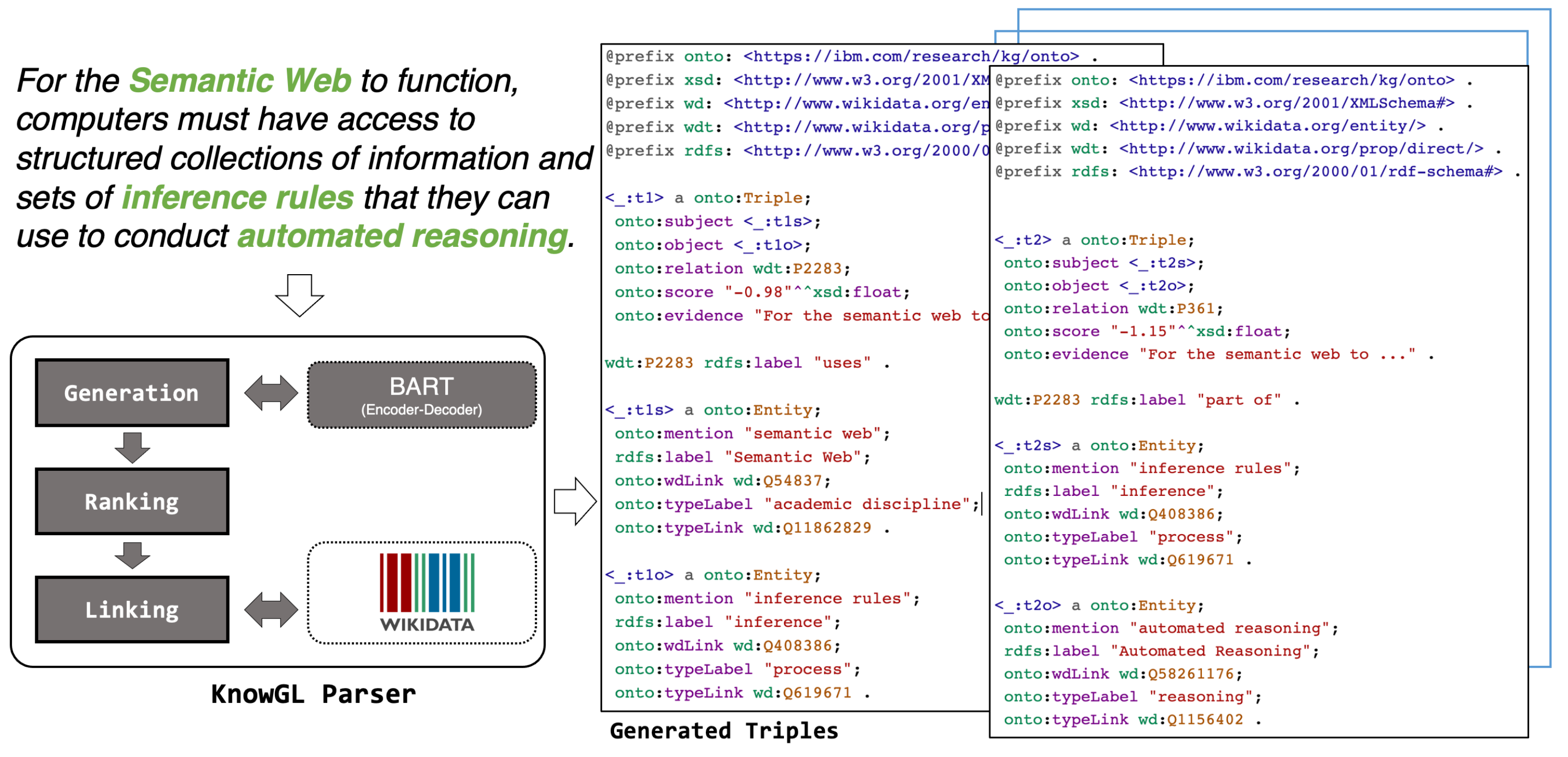}
\caption{The KnowGL Parser architecture and an example output.} \label{fig:ex_triple}
\end{figure}

A large part of our data is unstructured text. 
In order to incorporate them in the KG, we extract the textual values attached to each entity such as article content or a project description, chunk them into sentences and parse them using \emph{KnowGL Parser}, a novel tool integrated in our KGI framework~\cite{DBLP:conf/emnlp/GlassRCG21,DBLP:journals/corr/abs-2204-03985}.

KnowGL Parser allows converting unstructured text into structured data represented as a set of ABox assertions compliant with the TBox of Wikidata. 
We address this problem as a sequence generation task, similar to machine translation or text summarization, where the input is an English sentence and the output is a set of facts. To this aim we leveraged large pre-trained sequence-to-sequence language models, such as BART~\cite{DBLP:conf/acl/LewisLGGMLSZ20} and train them from large dataset derived using distant supervision, by exploiting the alignments between Wikidata facts with the abstracts of Wikipedia pages.

Specifically, given a sentence, we fine-tune the language model to detect pairs of entity mentions and jointly generate a set of facts (i.e. \textsc{<Subject (Subject Type), relation, Object (Object Type)>}) representing entity labels, entity types and their relationships. The output of the system is then deterministically converted in RDF statements, as shown in Figure~\ref{fig:ex_triple}. 

Our experiments and analysis show that KnowGL Parser produces more accurate triples improving in both precision and recall if compared with the state-of-the-art generative information extraction methods~\cite{DBLP:conf/semweb/RossielloMABGNK21,DBLP:conf/emnlp/CabotN21,DBLP:journals/corr/abs-2112-08340}.

Table~\ref{tab:wdparser_results} reports the F1 results of KnowGL Parser for each type of semantic annotations part of the triples generated from the abstracts, in terms of correct predictions of entity mentions, entity labels, entity types and their joint relations. For training and evaluation purposes, we extended a distantly supervised dataset for relation extraction~\cite{DBLP:conf/emnlp/CabotN21} with the full set of Wikidata-based annotations for each matched triple found in the abstracts of Wikipedia.

\begin{table}[h]
    \centering
    \begin{tabular}{r|cccc|ccc}
         & MD-F1 & TYPE-F1 & EL-F1 & RN-F1 & REL-P & REL-R & REL-F1  \\\hline \hline
         \multicolumn{7}{l}{\bf Approach} \\ \hline
         SOTA IE Pipeline~\cite{DBLP:journals/corr/abs-2112-08340} & - & - & - & - & 43.30 & 41.73 & 42.50 \\
         GenIE~\cite{DBLP:journals/corr/abs-2112-08340} & - & - & 79.69 & 78.21 & 68.02 & \textbf{69.87} & 68.93 \\\hline
         KnowGL Parser & 84.27 & 79.65 & \textbf{82.73} & \textbf{80.84} &  \textbf{73.88} & 67.85 & \textbf{70.74} \\ \hline \hline
    \end{tabular}
    \caption{Information extraction results. \textbf{MD} = Mention Detection. \textbf{TYPE} = Type Prediction. \textbf{EL} = Entity Label. \textbf{RN} = Relation Name. \textbf{REL} = Relation Prediction with Label Match. \textbf{P} = Precision. \textbf{R} = Recall. \textbf{F1} = Micro F1-score.}
    \label{tab:wdparser_results}
\end{table}


For both subject and object, we generate the surface form mention, canonical label, type label, relation label. Whenever applicable, we link the entities and types to Wikidata entities. Relations are also linked to Wikidata. This information is then converted in RDF and represented using a reified statement meta model. In addition, the facts are associated to an evidence attribute, which contains the provenance (i.e. the sentence) from which the triple has been generated together with its confidence score. An example output is shown in Figure~\ref{fig:ex_triple}. In addition, each triple is linked to the corresponding entity where the text was extracted.


\subsection{Implementation Details} 

The KG implementation consists of several components.  First and foremost is the actual deployment and hosting of the knowledge graph.  Our knowledge graph is hosted on a Blazegraph triplestore inside a RedHat OpenShift Container platform which gives us all the advantages of a cloud deployment (scaling, flexibility, storage).  We have a second component, a reverse proxy for Single Sign-On (SSO) authentication and authorization to the graph.  Some of the data in our graph is confidential and therefore requires a need to know access to prevent traversing and querying the graph by unintended parties.  The final set of components relate to the ETL (Extraction, Transform, and Load) process. Currently we build and load the graph on weekly basis. Our ETL process consists of extracting the data from all of the application APIs (both GraphQL and REST) as JSON documents, keeping an in-memory representation of the documents, and then converting these documents to RDF in Turtle format. The textual raw data of each entity is enriched with KnowGL Parser as described in Sec.~\ref{sec:wikidata_parser} with automated OpenShift cronjobs. Finally, RDF data coming from both structured  and textual sources is integrated and loaded into the triple store on a scheduled basis.

The current ETL process will be vastly be improved in the future to address the evolution of data by limiting text processing only to detected changes in the KG. Some of this future work will require including a text fingerprinting service to decide if the data has indeed changed (\textit{i.e.} for computational cost, we only care about the free text changes and not usually the meta data).

\section{Knowledge Exploitation Patterns}\label{sec:KEP}

To make the KG easy to use and adapt across different domains, we identified a set of common usage patterns, \emph{Knowledge Exploitation Patterns (KEP)}, and expose them as parameterized client API library to minimize the learning curve for the technology. These APIs generate the corresponding SPARQL queries and handle other cross-cutting concerns such as security or caching. Nevertheless, developers also can run queries directly in the SPARQL endpoint if needed. Currently, we provide APIs required for induced ontology exploration (type hierarchies, infoboxes), entity recommendation, and trend analysis. The idea behind the use of KEP is that certain functionalities can be abstracted out of the specific application domain by performing queries against the KG metamodel that is then used differently in downstream applications for the specific domain.

\subsection{Entity Recommendations}
Based on our use cases study described in Section~\ref{sec:use_case}, the automatic recommendations of items, such as publications, projects or collaborators, is one of the main desiderata for the members of our enterprise research community. 
Collaborative filtering~\cite{DBLP:reference/sp/KorenB15} is arguably the most common approach for recommendation systems, especially in environments with a large user base where the state-of-the-art methods are based on advanced deep learning techniques. 
However, an enterprise research community might not have enough users to train large parametric models due to the sparsity of user log activities.
For this reason, we adopt a hybrid content-based recommendation system method~\cite{DBLP:reference/sp/GemmisLMNS15,liu2021survey} by exploiting jointly the textual content, structured data and induced semantic annotations generated from our KnowGL Parser (see Section~\ref{sec:wikidata_parser}).

The idea is to convert our KG in an entity-feature Vector Space Model (VSM) model, where the rows are represented by the different type of entities in the KG, such as people, publications, projects and accomplishments, and the columns represent the feature space. 
In detail, let us consider $VSM^{n,m}$ a matrix using the standard tf-idf weighting schema, where each row $e_{i,*}$ is an entity vector created by concatenating different groups of features, described as follow: 

\begin{description}
\item[Bag of words] the textual content of entities, such as publications or projects, are tokenized and each token is considered as a single (sparse) feature. For entities representing people, where the textual context is not available, we exploit our KG to collect the textual content, e.g., from the publications or projects linked to the specific user by a multi-hop navigation in the graph.
\item[Structured data] this feature set represents relations derived from knowledge integration from our original data sources. For instance, the research division and topic of a project, the upper-line management for a person, and so on.
\item[Entities] this feature set represent the entities extracted from KnowGL Parser, grouped by their Wikidata type. For example, given the triples in Fig.~\ref{fig:ex_triple}, we create  entity features such as \textsc{Semantic Web:academic discipline}, \textsc{inference:process}, and so on.
\item[Frames] we also leverage the semantic relational information from the extracted triples. In order to alleviate the sparsity problem, we only concatenate the semantic annotations w.r.t. the domain, relation and range of each triple. For instance, \textsc{<academic discipline, uses, process>} for one of the generated triple in Fig.~\ref{fig:ex_triple}.
\end{description}

It is important to note that our feature set does not depend on the specific entity and relation set. Instead, this pattern is totally domain-agnostic and reusable and can be applied to any KG and entity type generated from our KG induction pipeline and integration process. 

After the VSM is built, the recommendation inference for a user is implemented in a non-parametric manner by exploiting the cosine similarity between the user and the target item vectors, such as publications, projects or other users. In other words, the recommended items for a user are the nearest neighbor entities in the vector space ranked by their cosine similarity scores.

\begin{figure}[h!]
\includegraphics[width=\textwidth]{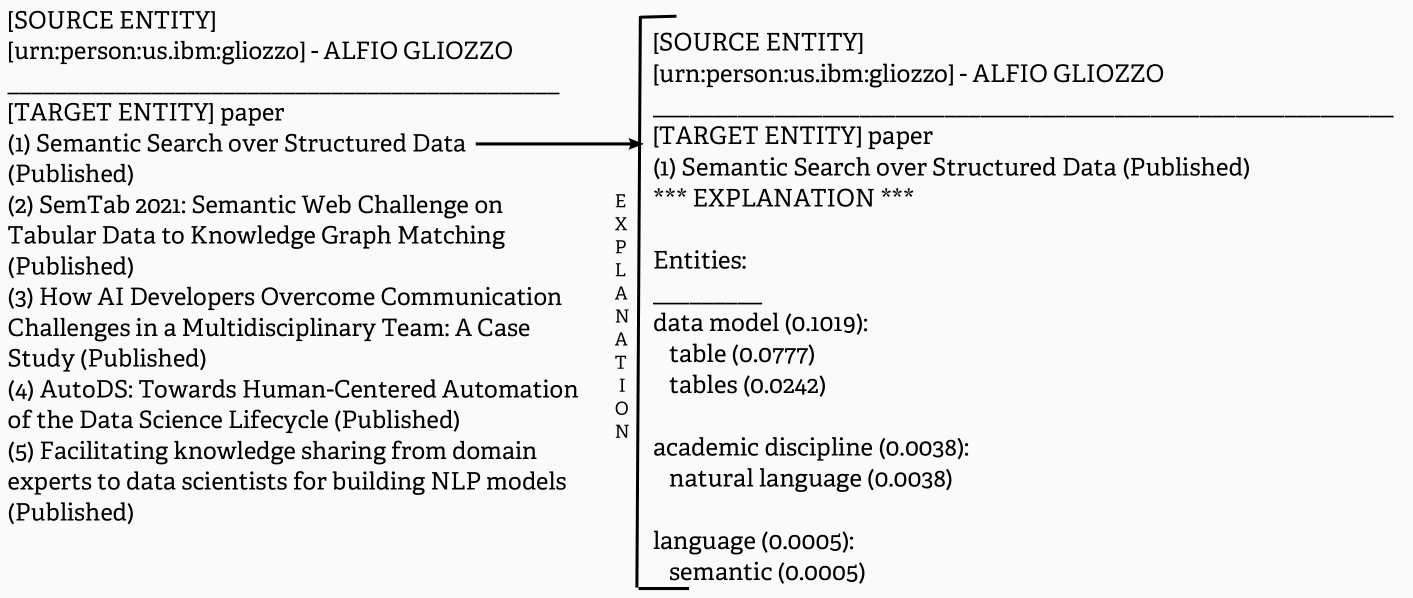}
\caption{An example of paper recommendations for a researcher. The figure on the left reports the list of recommended publications. The explanation for the top ranked item is shown in the figure on the right as a list of relevant entities grouped by their semantic types.}
\label{fig:rec_example}
\end{figure}

Figure~\ref{fig:rec_example} shows an example of a list of recommended publications for an researcher using the aforementioned KG-based VSM. The KG induced from text allows us to provide meaningful explanations for the user that justify the recommendation. The explanation is obtained by measuring and selecting the most relevant entities (i.e. those that contributed most to the similarity score), ranked by their combined tf-idf weights.

To evaluate the quality of the recommendations, we recruited 30 volunteer researchers from various disciplines. For each participant, we recommended 10 projects, 10 papers and 5 achievements. Each participant was asked to rate the recommendations on the following scale:

\begin{itemize}
    \item NONE: No value to me
    \item LOW: Good to know but I am not going to read anytime soon
    \item MEDIUM: Relevant for my specific area of interest (must read)
    \item HIGH: Relevant to my current project(s) and work
\end{itemize}


We performed a quantitative analysis by evaluating Mean Average Precision (MAP) and Precision@K (P@K) metrics,  which are popular choices to evaluate recommendation systems. 
Both MAP and P@K take in consideration only binary assessments, i.e. if the recommended item is relevant or non-relevant.
In order to convert our graded rating into a binary assessment, we adopt three different criteria, namely HIGH (i.e. only HIGH category is regarded as positive), MEDIUM (i.e. HIGH and MED categories are positive), LOW (i.e. HIGH, MED and LOW are positive).
As shown in Table~\ref{tab:rec_results}, the performance of our recommendation system is consistent across the different type of recommended items. Moreover, the MAP is consistently higher than P@K, showing that the system tends to provide higher scores to those items considered relevant for the users.

\begin{table}[t]
\centering
\bgroup
\setlength{\tabcolsep}{5pt}
\def\arraystretch{1.2}
\begin{tabular}{l|cc|cc|cc} 
\toprule
\multicolumn{1}{c|}{\textbf{}} & \multicolumn{2}{c|}{\centering \textbf{Papers}} & \multicolumn{2}{c|}{\centering \textbf{Projects}} & \multicolumn{2}{c}{\centering \textbf{Achievements}} \\ \hline
Criteria& MAP &P @ 10 & MAP & P @ 10 & MAP & P @ 5\\ \hline
\textbf{LOW} & 0.89 & 0.76 & 0.92  & 0.81 & 0.87 & 0.82   \\
\textbf{MEDIUM} &  0.51 & 0.34 & 0.65  & 0.45 & 0.51 & 0.36    \\
\textbf{HIGH} &  0.21 & 0.08 & 0.50  & 0.14 & 0.41 & 0.17  \\ 
\bottomrule
\end{tabular}
\egroup
\caption{User evaluation for scholary article, project and achievement recommendations for 30 users.}
\label{tab:rec_results}
\end{table}

We also performed an analysis focusing on irrelevant recommendations. One repeating pattern was the users who have recently moved to a different research area tends to have less accurate recommendation. This is can be explained by observing that their historical publication profile did not reflect their current information needs.
Another commonly reported problem is that in many cases the researchers were aware of the recommended items already, in spite of the fact that we filtered out those items were they were explicitly listed as authors or contributors. The explanation for that is that there could be multiple relations between a person and an information object, besides being \texttt{authorOf}. For example, one researcher might have been the mentor of one of those authors, might have been part of a review committee and so on. In future work, we planned to address the above issues by applying more sophisticated machine learning-based recommendation techniques able to learn how to traverse the graph structure from the user provided feedback.


\subsection{Trend Analysis}

The KG induced from the unstructured text is used to implement KEPs for trend analysis. Once a corpus is completely processed by KnowGL Parser, trend analytics provide an overview of the concepts found in the corpus simply by performing aggregation queries the induced KG. 

Since we can not show examples of analytic from the IBM internal data due to privacy of strategic information, we have created a KG by processing ISWC papers from year 2002-2021 using DBLP RDF data\footnote{\url{https://blog.dblp.org/2022/03/02/dblp-in-rdf/}}. For each paper, we collected the title and the abstract of the paper and parsed them using the KnowGL Parser to create an Induced KG. Examples in this section uses that KG. This also provides evidence that the approach that we have proposed can be easily adapted to other communities. 

\begin{figure}[h!]
\includegraphics[width=\textwidth]{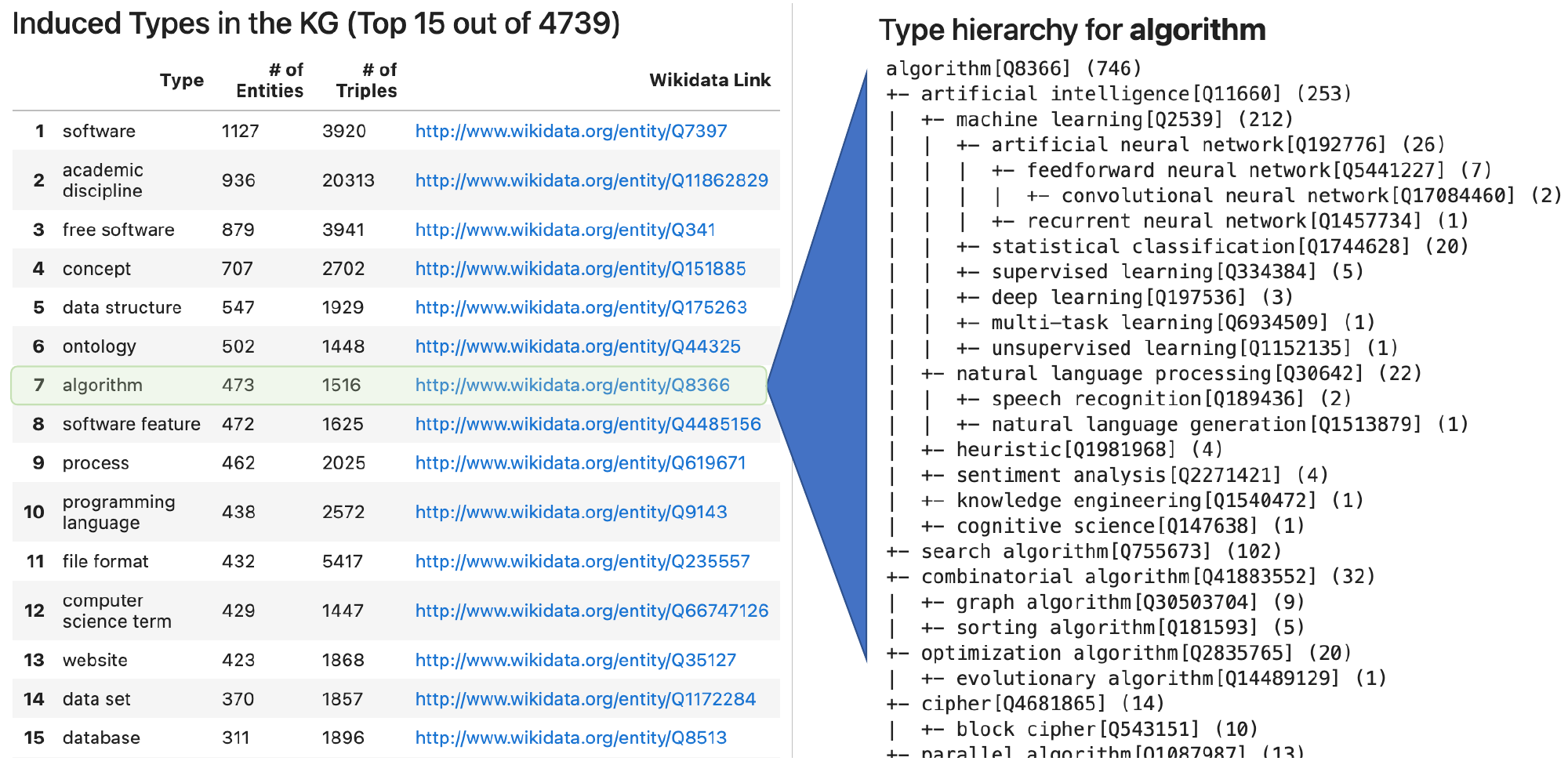}
\caption{A snippet from induced types from the ISWC corpus.} \label{fig:types}
\end{figure}

Fig.~\ref{fig:types}  shows the most frequent types found in the ISWC 2002-2021 corpus with the number of unique entities found in the corpus and number of associated triples. Any type can be selected and expanded to see its subtypes in the corpus ordered by their cumulative frequency (direct children and all descendants). Fig.~\ref{fig:types} (right) illustrates the expansion of type \texttt{algorithm} which has 473 direct entities and 746 transitive entities. The subclass relations are both induced from text and extracted from Wikidata. Users can select any of 4739 types generated in the case of the ISWC corpus and generate a trend analysis for the given type.

\begin{figure}[h!]
\includegraphics[width=\textwidth]{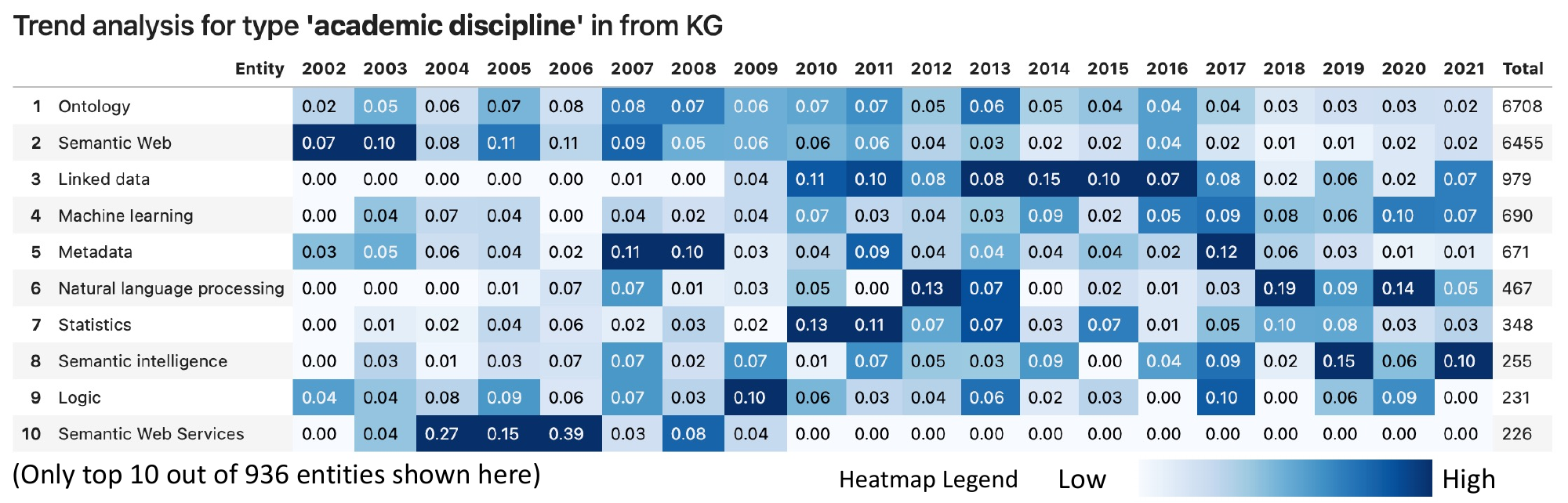}
\caption{An example of trends analytics for entities of type academic discipline} \label{fig:trends}
\end{figure}

Figure~\ref{fig:trends} shows the trend analysis for entities belonging to the type \texttt{academic discipline}. The last column shows the total number of occurrences of each entity in all ISWC papers from 2002 - 2021. Individual cells show the distribution of the papers in different years as a percentage. Such trend analysis can highlight some interesting facts. For instance, it shows that there was a high interest in ``\textit{Ontology}" and ``Semantic Web" throughout from the beginning but the interest diversify more in later years. Similarly, we can see that there is a high interest in ``\textit{Linked Data}" from year 2009 which is at highest during the 2013 - 2017 period. In contrast, ``\textit{Semantic Web Services}" are of high interest during 2003 - 2009 period but the interest completely vanishes on later years. It is important to notice that the list of entities belonging to the type \texttt{academic discipline} or any other type is automatically generated. The analyst is just supposed to point to the right concept in the taxonomy to get her job done.

\subsection{Infobox Generation}

\begin{figure}[h!]
\includegraphics[width=\textwidth]{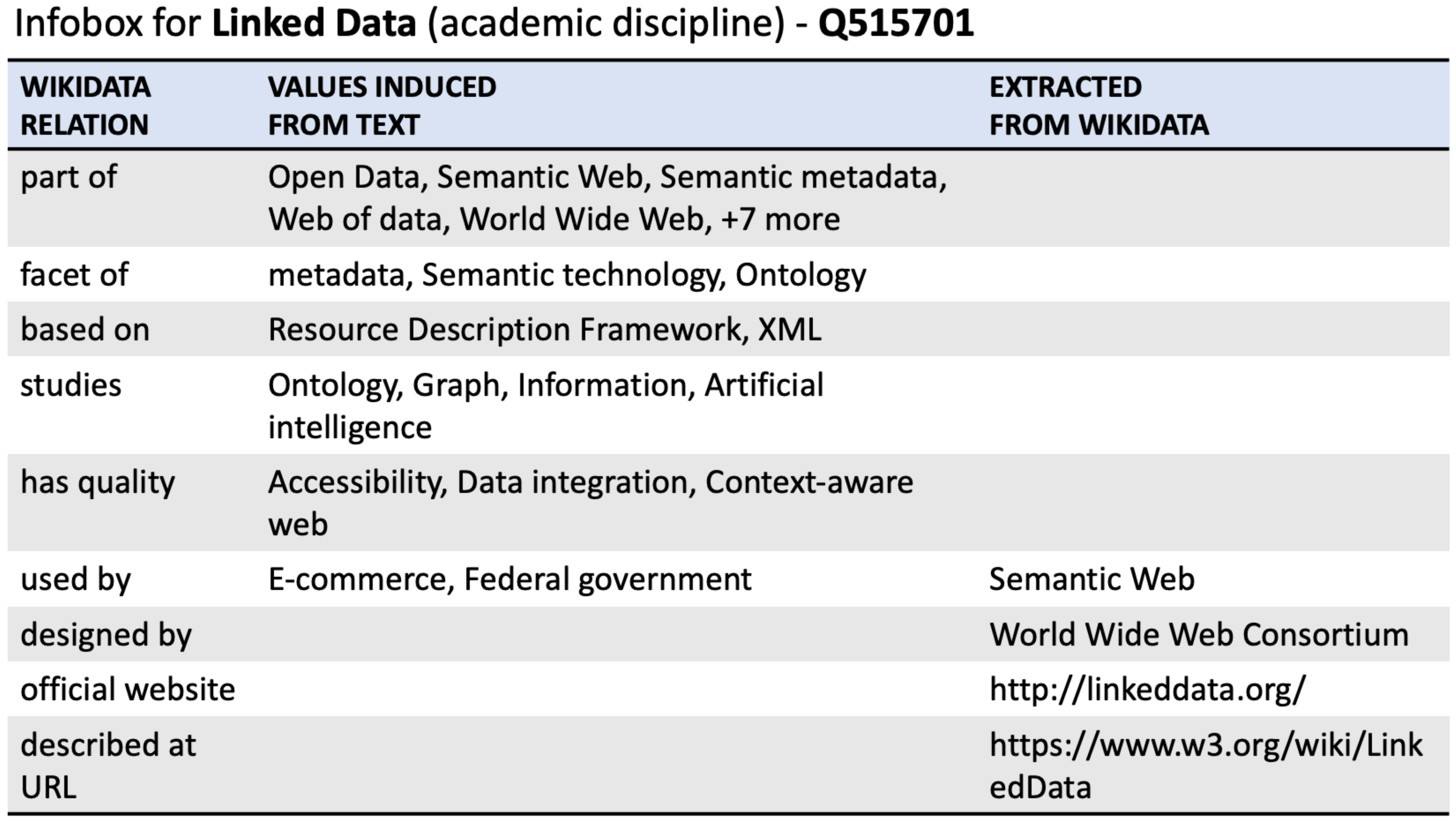}
\caption{An example of infobox for ``\textit{Linked Data}" an entity including both induced facts and integrated Wikidata facts.} \label{fig:infobox}
\end{figure}

\begin{figure}[h!]
\includegraphics[width=\textwidth]{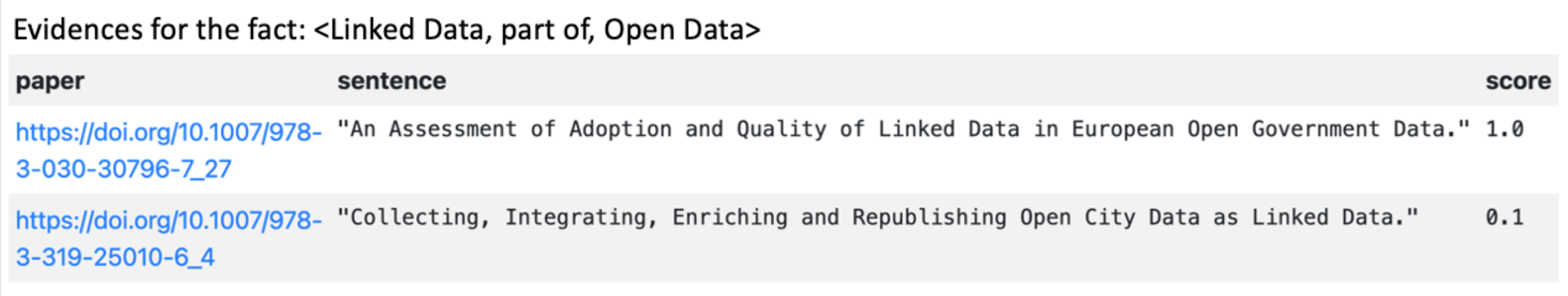}
\caption{Evidences for the induced fact (Linked Data, part of, Open Data)} \label{fig:evidence}
\end{figure}

Once an entity of interest is selected, for example, ``\textit{Linked Data}", the users can automatically generate an infobox, as shown in Fig.\ref{fig:infobox}. We first induce a schema for each type, by counting the most most frequent relations extracted by the parser for entities of that type. For example, for the type \texttt{academic discipline} the important relations are \texttt{part of}, \texttt{facet of}, \texttt{based on}, \texttt{studies} and so on. Then we collect the object filling those relations for a specific target entity (\textit{Linked Data}, in the example). Those relations might come from induced triples or from Wikidata itself. Each of the relations in the infobox is also associated to its provenance (might be a textual occurrence or a pre-existing triple in Wikidata) as illustrated by Fig.~\ref{fig:evidence}. 

\section{Related Work}\label{sec:relatedwork}

KGs are a common way to organize data from multiple sources providing a unified view and represent them in a semantically rich manner empowering a wide range of downstream applications~\cite{hogan2021knowledge, ji2021survey, noy2019industry}. More specifically, Scholarly KGs such as ORKG~\cite{jaradeh2019open}, MAG~\cite{wang2020microsoft}, OpenAIRE~\cite{manghi2012data} are becoming popular way to represent research data. Such KGs are used for search~\cite{heidari2021demonstration, auer2021leveraging}, question answering~\cite{jaradeh2020question}, recommendation~\cite{nayyeri2020embedding, manrique2018knowledge}, analysis of research trends~\cite{salatino2021detection}, performing surveys~\cite{oelen2020generate}, and understanding the dynamics between academia and industry~\cite{angioni2020integrating}. 

The Semantic Web community has developed several methods and tools for building KGs. There are comprehensive survey articles on building KGs from relational databases~\cite{sahoo2009survey}, semi-structued data~\cite{ryen2022building}, and unstructured text~\cite{al2020automatic,DBLP:conf/aaai/CarlsonBKSHM10,DBLP:conf/vlds/NiuZRS12,DBLP:conf/kdd/0001GHHLMSSZ14,DBLP:conf/emnlp/MesquitaCSMB19}. Rezayi et al.\cite{edge_naacl21} propose an approach to augment a KG with key phrases generated from textual content of entities. In our work, we augment our KG with semantically rich triples generated from textual content of each entity. Furthermore, we integrate the induced knowledge with the relevant portion of background knowledge from Wikidata. 

Trend analysis on KGs has been used for analysing research topics~\cite{salatino2021detection, kim2017practical, wohlgenannt2014using}, patents\cite{weber2013ontology}, market trends~\cite{albrecht2019business}. KnowGL Parser presented in our approach allows automatically create an induced knowledge graphs from text with a large number of Wikidata types (~50K in 2022) enabling fine-grain analysis and seamless integration of background knowledge from Wikidata that can be used in the analysis.

Cai et al.~\cite{cai2022explicable} proposes an explainable recommender by generating the candidates using a KG and using an evolutionary algorithm. We use a simpler vector space model to produce recommendations between different types of entities.


\section{Conclusions and Future Work}\label{sec:conclusion}

In this work, we presented an application of the Knowledge Graph Induction (KGI) technology to fulfill the requirements identified by a user study to enhancing cooperation in a research community. We have shown how the induced knowledge enables several downstream applications, such as recommending and trend analytics, providing evaluation for most of the component based on both quantitative and qualitative approaches. 
This year, we intend to deploy the recommending technology to all the member of the IBM research community, in the order of 6,000 people. We envision both in-app and ``meet users where they are'' experiences outside the apps. In all cases, we will provide feedback mechanisms (e.g. thumbs up/down, free text explanations) for users to share their view on the quality of the recommendations. The intention is to feed this back into a deep learning based recommender to learn how to better exploit the graph traversals. 




In addition to trend analysis, we believe that KGI technology could also be leveraged for flexible and on-demand business analytics, providing powerful insights to accelerate business, for example:
\begin{description}
\item[Predicting success] What are the characteristics of research projects that result in recognition and awards. How do we invest in new projects that exhibit these characteristics to better steer the IBM research agenda? Which papers should we support to have the best chance of publication at key conferences?
\item[Business development] Quickly identifying relevant research activity of interest to current or prospective clients or partners
\item[Operations and efficiency]  Who is working on what projects and is time being used effectively? Is there duplicate activity? Where are the gaps? What are best opportunities for cross-collaboration?
\item[Talent] Who are the rising stars? How do we find the right projects for them, or nominate them for external awards? 
\item[Portfolio] Tracing research projects and outcomes to Objects and Key Results.
\end{description}

We plan to develop KEP for the use cases above that can be generalized beyond the research community use case. We believe that the KEPs can be designed to cover variety of different use cases in many different organizations. 

Moreover, we are planning to acquire KGs from different research communities (e.g. Semantic Web, NLP, Deep Learning communities) and make them available to the community. The goal is to act as a catalyzer for future research work in the research community beyond IBM.

\bibliographystyle{splncs04}
\bibliography{iswc2022}

\end{document}